\documentclass[10pt,twocolumn,letterpaper]{article}

\makeatletter
\@namedef{ver@everyshi.sty}{}
\makeatother
\usepackage{iccv}
\usepackage{times}
\usepackage{epsfig}
\usepackage{graphicx}
\usepackage{amsmath}
\usepackage{amssymb}
\usepackage{svg}

\usepackage[breaklinks=true,bookmarks=false]{hyperref}

\iccvfinalcopy 

\def\iccvPaperID{****} 

\ificcvfinal\fi

\begin{document}

\title{MobileStyleGAN: A Lightweight Convolutional Neural Network for High-Fidelity Image Synthesis}

\author{Sergei Belousov\\
{\tt\small sergei.o.belousov@gmail.com}
}

\maketitle
\ificcvfinal\thispagestyle{empty}\fi

\begin{abstract}
    In recent years, the use of Generative Adversarial Networks (GANs) has become very popular in generative image modeling. While style-based GAN architectures yield state-of-the-art results in high-fidelity image synthesis, computationally, they are highly complex. In our work, we focus on the performance optimization of style-based generative models. We analyze the most computationally hard parts of StyleGAN2, and propose changes in the generator network to make it possible to deploy style-based generative networks in the edge devices. We introduce MobileStyleGAN architecture, which has x3.5 fewer parameters and is x9.5 less computationally complex than StyleGAN2, while providing comparable quality.
\end{abstract}

\section{Introduction} \label{introduction}

\begin{figure*}
    \begin{center}
    \includegraphics[scale=0.1]{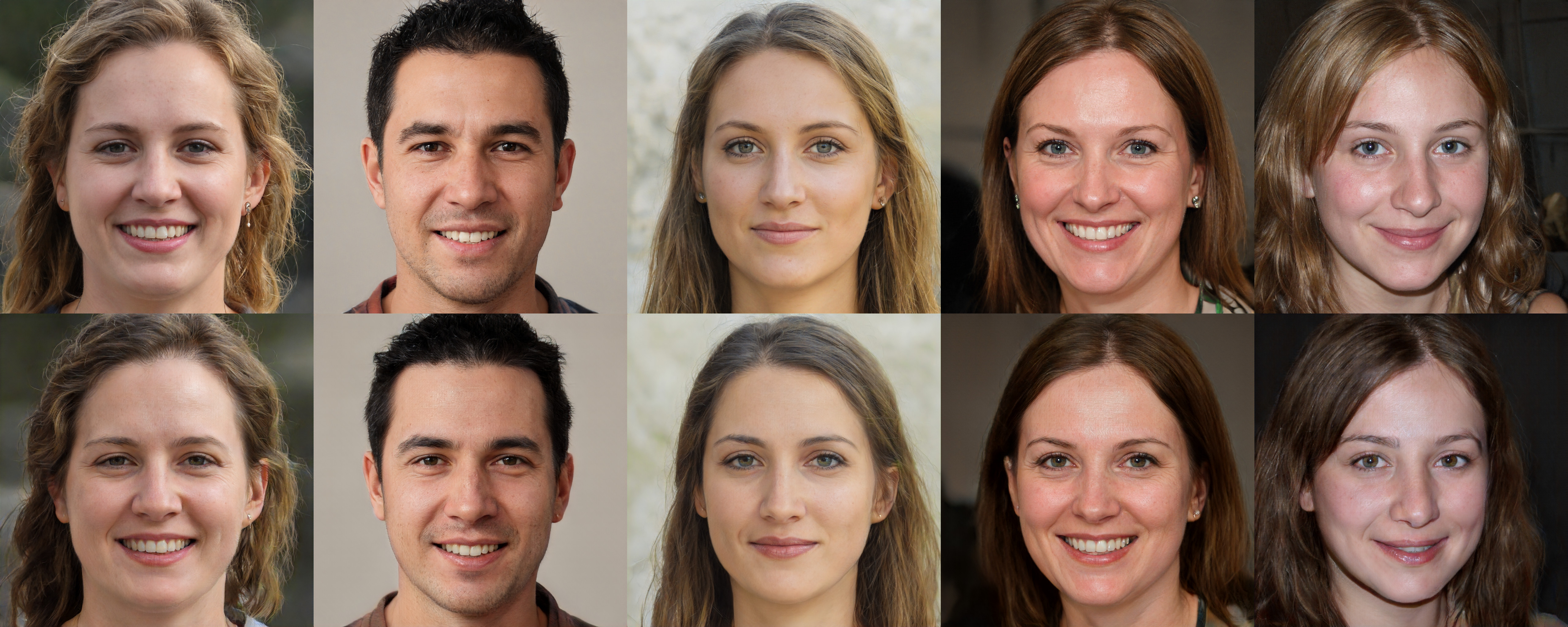}
    \end{center}
    \caption{(top) Images generated by StyleGAN2. (bottom) Images generated by MobileStyleGAN.}
    \label{fig:stylegan_mstylegan}
\end{figure*}

In recent years, high-fidelity image synthesis has significantly improved by through the use of Generative Adversarial Networks (GANs) \cite{goodfellow2014generative}. Whereas early work such as DCGAN \cite{radford2016unsupervised} could generate images having a resolution up to 64x64 pixels, modern networks such as BigGAN \cite{brock2019large} and StyleGAN \cite{karras2019stylebased, Karras2019stylegan2, Karras2020ada}  allow the generation of photorealistic images with up to 512x512 and even 1024x1024 pixels. Although the quality of generative models has significantly improved, image generation still requires many computation resources. The high computational complexity makes it difficult to deploy state-of-the-art generative models to edge devices.

For example, the StyleGAN2 \cite{Karras2019stylegan2} network allows realistic face images 1024x1024 pixels in size with FID=2.84 for the FFHQ dataset. It, however, contains 28.27M parameters and has a computational complexity of 143.15GMAC.

We propose a new lightweight architecture, MobileStyleGAN, a high-resolution generative model for high-quality image generation. Taking as a baseline the original StyleGAN2 architecture, we revisit computationally hard parts of this network to create our own lightweight model that provides comparable quality (Figure \ref{fig:stylegan_mstylegan}). The whole network contains 8.01M parameters, has a computational complexity of 15.09 GMAC, and provides quality with FID=7.75 for the FFHQ dataset.

Our main contributions are:
\begin{itemize}
    \item We introduce an end-to-end wavelet-based convolutional neural network for high-fidelity image synthesis.
    \item We introduce Depthwise Separable Modulated Convolution as a lightweight version of Modulated Convolution to decrease computational complexity.
    \item We introduce a revisited version of the demodulation mechanism applicable to graph optimizations such as operation fusion.
    \item We propose a pipeline based on knowledge distillation to train our network.
\end{itemize}

\section{Related Work}
\subsection{StyleGAN}
StyleGAN \cite{karras2019stylebased} is a modern generative model for high-resolution image generation. The key aspects of the StyleGAN network are:
\begin{itemize}
    \item It uses progressive growing to increase the resolution gradually.
    \item It generates images from a fixed value tensor, as opposed to generating images from stochastically generated latent variables as in conventional GANs.
    \item The stochastically generated latent variables are used as style vectors through AdaIN \cite{huang2017arbitrary} at each resolution after being nonlinearly transformed by an 8-layer neural network.
\end{itemize}

StyleGAN2 \cite{Karras2019stylegan2} improves upon StyleGAN by:
\begin{itemize}
    \item Eliminating droplet modes by normalizing with estimated statistics instead of normalizing with actual statistics such as AdaIN.
    \item Reducing eye and tooth stagnation by using a hierarchical generator with skip connections instead of progressive growing.
    \item Improving image quality by reducing PPL and smoothing the latent space.
\end{itemize}

StyleGAN2-ADA \cite{Karras2020ada} makes StyleGAN applicable to tasks with limited data by using adaptive discriminator augmentations.

\subsection{Model acceleration}

A prominent research directions in deep learning focuses on accelerating to convolution neural networks (CNNs) using manual and automatic design of lightweight architectures.

Some works focused on designing efficient neural classification networks that can be used as a backbone for other tasks. Howard et al. \cite{howard2017mobilenets, howard2019searching, sandler2019mobilenetv2} proposed efficient models called MobileNets for mobile and embedded vision applications. These ideas have been improved in many articles \cite{tan2020efficientnet, iandola2016squeezenet, gholami2018squeezenext, zhang2017shufflenet, ma2018shufflenet}.

Other works focused on the design models of efficient generative models. Li et al. \cite{li2020gan} proposed a generic framework for automatic optimization of the conditional GANs based on a distillation and neural architecture search \cite{cai2020onceforall}. Chang and Lu \cite{chang2020tinygan} proposed a distillation pipeline for compression of BigGAN. Liu et al. \cite{liu2021faster} proposed a lightweight GAN based on the attention mechanism.

\subsection{Knowledge distillation}

Hinton et al. \cite{hinton2015distilling} proposed the knowledge distillation method to train a small student network using a large teacher network. The main idea of the distillation is that the student network is trained to mimic the teacher network's behavior. Aguinaldo et al. \cite{aguinaldo2019compressing} adopted knowledge distillation to accelerate unconditional GANs. Some other works related to GANs \cite{wang2020gan, fu2020autogandistiller, li2020gan, chang2020tinygan} also used knowledge distillation as part of their pipeline.

\subsection{Wavelet transform}
Using wavelet-based methods with deep learning is not novel. Wavelets \cite{HaarOnTT} have been applied to many computer vision tasks such as texture classification \cite{fujieda2017wavelet}, image restoration \cite{liu2018multilevel},and super-resolution \cite{zhang2019image}.

Han et al. \cite{han2020notsobiggan} proposed Not-So-Big-GAN architecture based on two sub-networks: a generative network in low resolution and a super-resolution network for upsampling. The authors showed that wavelet-based sub-networks outperform pixel-based methods.

In comparison to previous works, we present an end-to-end wavelet-based CNN architecture for generative networks. We show that integrating wavelet-based methods into GANs allows more lightweight networks to be designed and provides more smoothed latent space.

\section{MobileStyleGAN Architecture}
Our proposed architecture, MobileStyleGAN, is based on previous works related to style-based generative models. MobileStyleGAN includes a mapping network and synthesis network, as in StyleGAN2. We adopt the mapping network from StyleGAN2, focusing on the design of a computationally efficient synthesis network.

We now first describe the key differences between our proposed architecture and the basic StyleGAN2. We then describe the distillation-based training procedure for the MobileStyleGAN network.

\subsection{Image representation revisited} \label{image_representation_revisited}

While StyleGAN2 works with pixel-based image representation and aims to directly predict the pixel values of the output image, in our work, we use a frequency-based representation of images. In this way, MobileStyleGAN seeks to predict the discrete wavelet transform (DWT) of the output image.

\begin{figure}[t]
\begin{center}
   \includegraphics[width=1.0\linewidth]{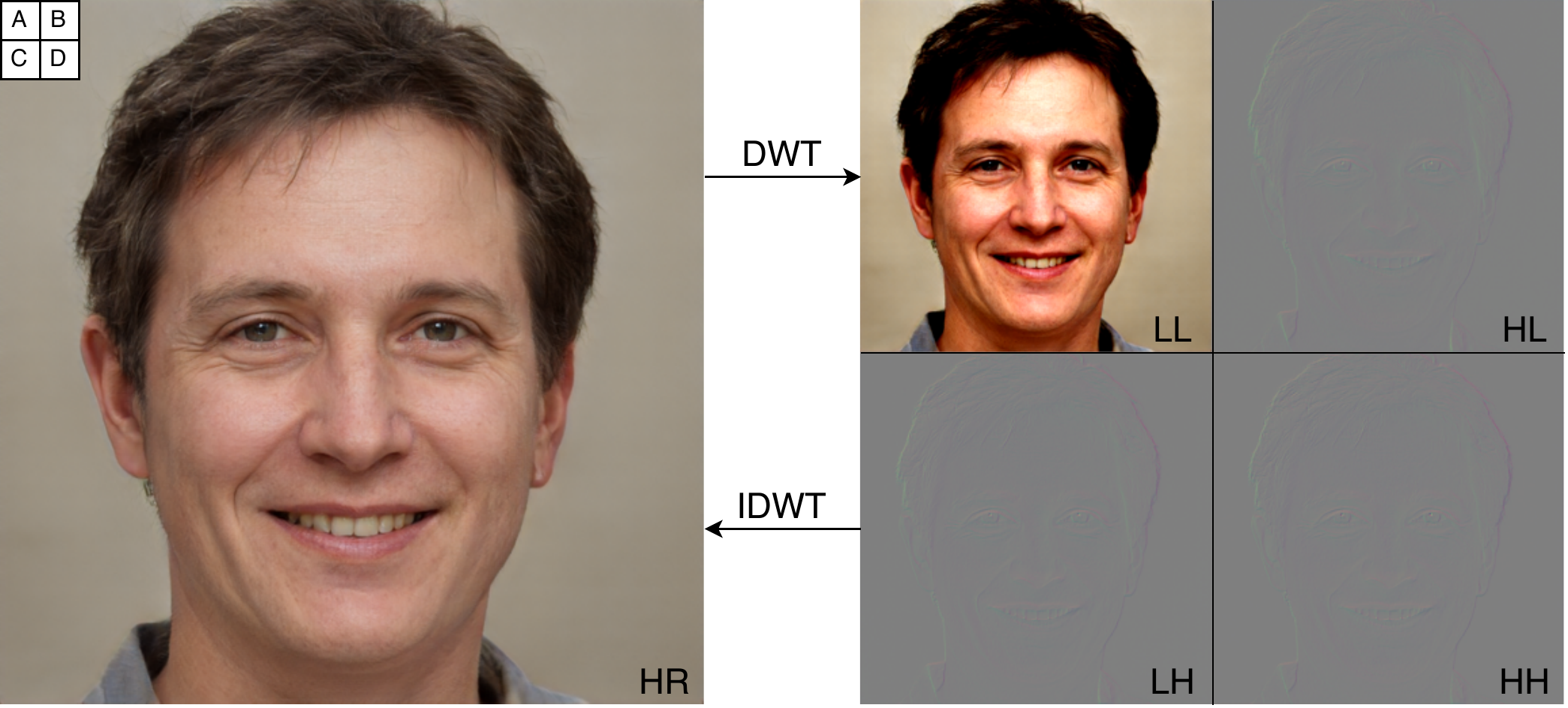}
\end{center}
   \caption{(left) The original RGB image. (right) DWT decomposition of the image. A, B, C, D are four pixels located in the top-left grid 2x2 of the HR image.}
\label{fig:dwt_idwt}
\end{figure}

When applied to a 2d image, the DWT transforms the channel into four equally-sized channels having a lower spatial resolution and different frequency bands. The inverse discrete wavelet transform (IDWT) then reconstructs the pixel-based representation from the wavelet-domain (Figure \ref{fig:dwt_idwt}).

This type of representation of images has several advantages, such as:
\begin{itemize}
    \item Since wavelet-based image representation contains more structural information than pixel-based approaches, we can generate high-resolution images using low-resolution feature maps without losing accuracy:
    \begin{equation}
    \begin{array}{l}
        |I - IDWT(DWT(I))|_{l1} < \epsilon \\
        \epsilon \approx 1e-7
    \end{array}
    \end{equation}

    \item In our work, we use Haar wavelets \cite{HaarOnTT} as a bank of filters for the DWT and IDWT. IDWT using Haar wavelets can be implemented efficiently without multiplication operations (Figure \ref{fig:dwt_idwt}):
    \begin{equation}
    \begin{array}{l}
    \left\{\begin{matrix}
    A = LL + HL + LH + HH\\
    B = LL - HL + LH - HH\\
    C = LL + HL - LH - HH\\
    D = LL - HL - LH + HH
    \end{matrix}\right.
    \end{array}
    \end{equation}

    \item The generation of the high-frequency details of the image is a complex problem. While the latent space of StyleGAN is smoothed in low frequencies, it is rough in high frequencies.  In contrast to pixel-based approaches, using frequency-based image representation, we can add regularization to the high-frequency components of the signal directly, which makes latent space smooth at both low and high frequencies.
\end{itemize}

\subsection{Progressive growing revisited} \label{progressive_growing_revisited}
StyleGAN2 uses a skip-generator to form the output image by explicitly summing RGB values from multiple resolutions of the same image. We found that when we predict images in the wavelet domain, the prediction head based on skip connections does not make an essential contribution to the quality of the generated image. Accordingly, to decrease computational complexity, we replace the skip-generator with a single prediction head from the network's last block. Predicting the target image from the intermediate blocks, however, is important in stabilizing image synthesis. We, therefore, add an auxiliary prediction head for each intermediate block to predict the target image according to its spatial resolution.

\begin{figure}[t]
\begin{center}
   \includegraphics[width=1.0\linewidth]{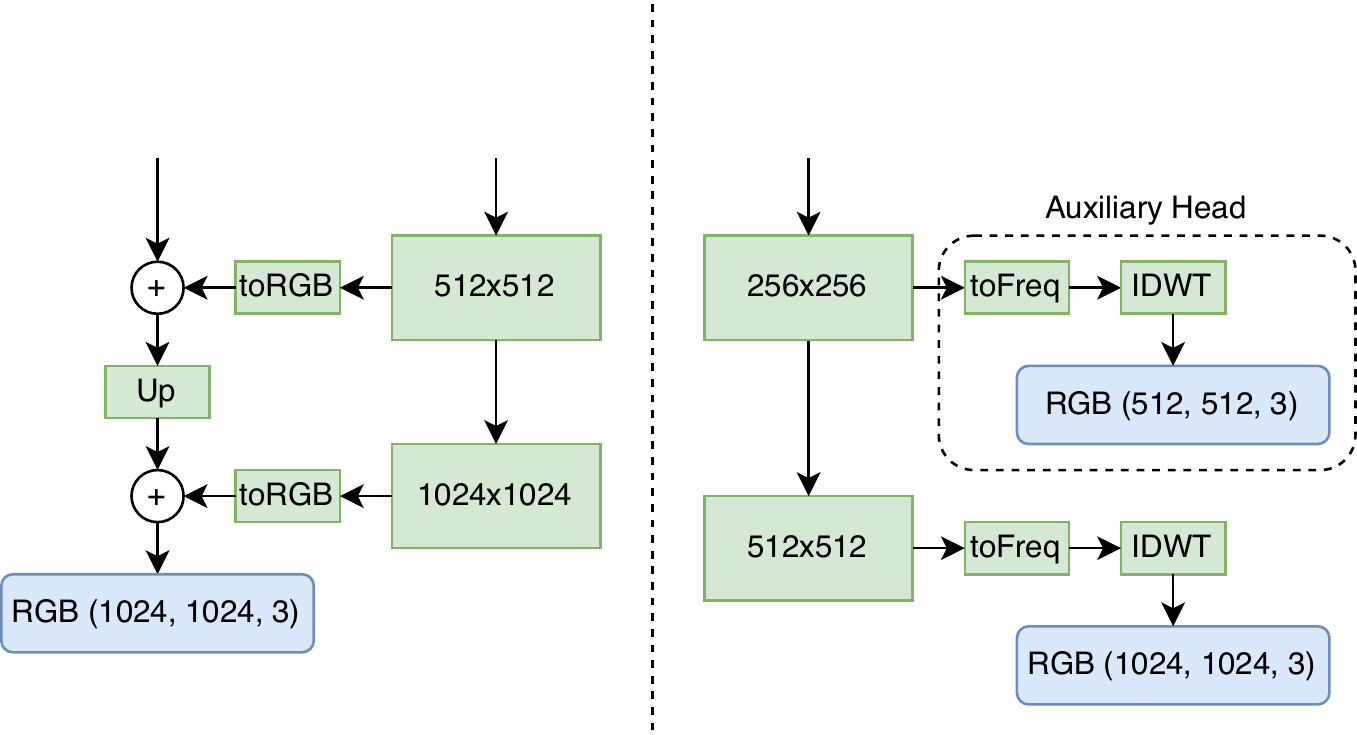}
\end{center}
   \caption{(left) Prediction head of StyleGAN2. (right) Prediction head of MobileStyleGAN.}
\label{fig:prediction_head}
\end{figure}

The difference between the StyleGAN2 and MobileStyleGAN prediction heads is shown in Figure \ref{fig:prediction_head}.

\subsection{Depthwise separable modulated convolution}

\begin{figure*}
    \begin{center}
    \includegraphics[width=1.0\linewidth]{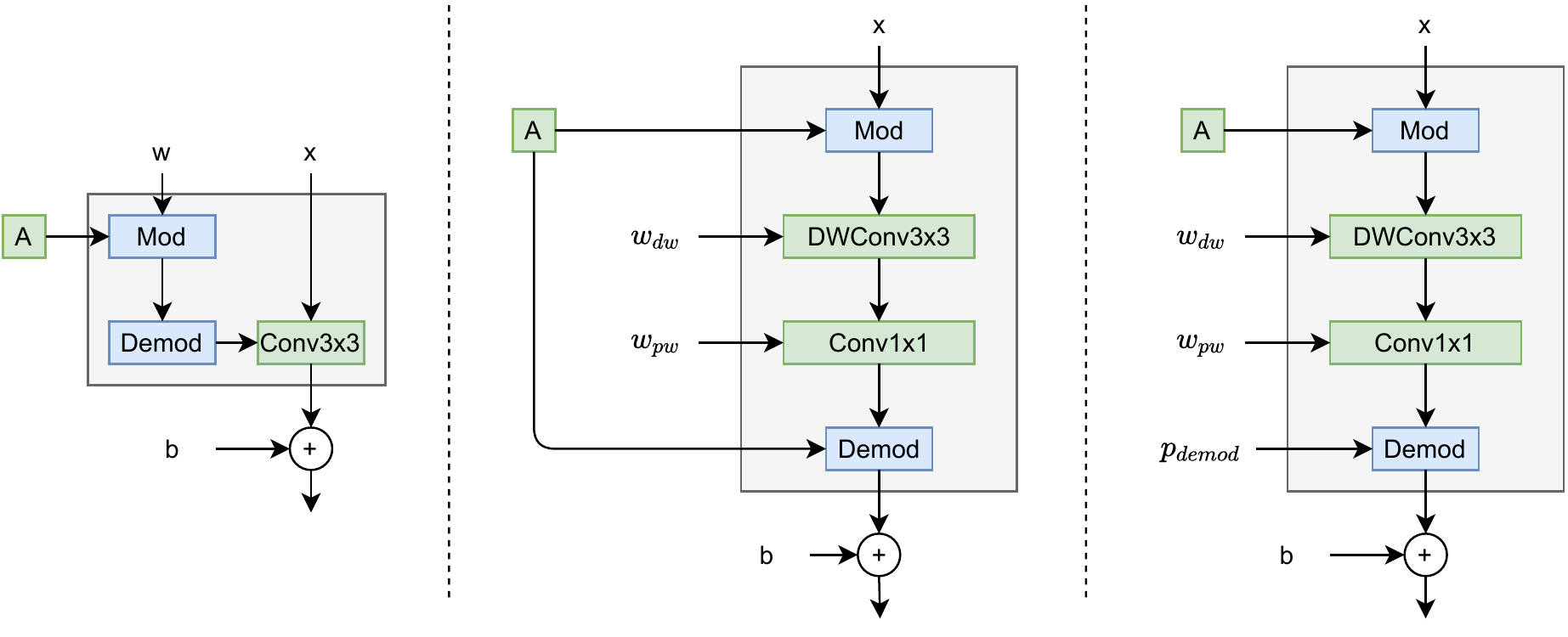}
    \end{center}
    \caption{(left) Modulated convolution. (middle) Depthwise separable modulated convolution. (right) Depthwise separable modulated convolution with a trainable demodulation.}
    \label{fig:dw_demodulated_convolution}
\end{figure*}

Inspired by MobileNet \cite{howard2017mobilenets}, MobileStyleGAN is based on depthwise separable convolutions, a form of factorized convolutions that factorize a standard convolution into a 3x3 depthwise convolution and a 1x1 convolution called a pointwise convolution.

As described in \cite{Karras2019stylegan2}, a modulated convolution consists of modulation, convolution, and normalization (left panel of Figure \ref{fig:dw_demodulated_convolution}). Depthwise separable modulated convolution also comprises these parts (middle panel of Figure \ref{fig:dw_demodulated_convolution}). We note, however, that while StyleGAN2 describes modulation/demodulation applied to weights, we apply them to input/output activations, respectively. This order of operations makes it easy to describe depthwise separable modulated convolution. Let us start by considering the effect of a modulation followed by a convolution. The modulation scales each input feature map of the convolution based on the incoming style:
\begin{equation}
    x'=s*x
\end{equation}
where $x$ and $x'$ are the original and modulated input activations, respectively, and $s$ is the scale corresponding to the input feature maps. Then we apply 3x3 depthwise convolution and 1x1 pointwise convolution sequentially without any nonlinearity between them:
\begin{equation}
    \begin{array}{l}
    x'' = w_{dw} * x' \\
    x''' = w_{pw} * x''
    \end{array}
\end{equation}
Now we apply demodulation to remove the effect of $s$ from the statistics of the output feature maps. Due to the linearity of the convolution operator, the result of the sequentially applied depthwise and pixelwise convolutions are equal to the result of the applied dense convolution:
\begin{equation}
    \begin{array}{l}
    w_{dense} = w_{dw} * w_{pw} \\
    w_{pw} * (w_{dw} * x') = (w_{pw} * w_{dw}) * x' = w_{dense} * x'
    \end{array}
\end{equation}
In this way, demodulation coefficients can be calculated as:
\begin{equation}
    \begin{array}{l}
    w' = w_{pw} * w_{dw} \\
    demod_{j} = \frac{1}{\sqrt{\sum_{i, k}{(s_i * w'_{i,j,k})^{2}} + \epsilon}}
    \end{array}
    \label{demod_eq}
\end{equation}
where $i$ and $j$ and $k$ enumerate the input/output feature maps and spatial footprint of the convolution, respectively. To demodulate the output feature maps, we apply:
\begin{equation}
    x_{out} = demod * x'''
\end{equation}

\subsection{Demodulation fusion}

Batch normalization fusion is a popular technique to decrease the computational complexity of convolutional networks at the inference time. This technique relies on the fact that we can merge two linear operations into one. The demodulation mechanism is similar to batch normalization, but it is not a linear operation at the inference time. Following Equation \ref{demod_eq}, where weights are fixed, the demodulation is the function of style. To make the demodulation constant, we replace style coefficients with trainable parameters (right panel of Figure \ref{fig:dw_demodulated_convolution}):
\begin{equation}
    \begin{array}{l}
    w' = w_{pw} * w_{dw} \\
    demod_{j} = \frac{1}{\sqrt{\sum_{i, k}{(p_{demod} * w'_{i,j,k})^{2}} + \epsilon}}
    \end{array}
\end{equation}
Thus, demodulation becomes the constant at the inference time and can be merged into the pixelwise convolution weights. We found that this technique does not adversely affect the quality of the generated image.

\subsection{Upscale revisited}

\begin{figure}[t]
    \begin{center}
    \includegraphics[width=1.0\linewidth]{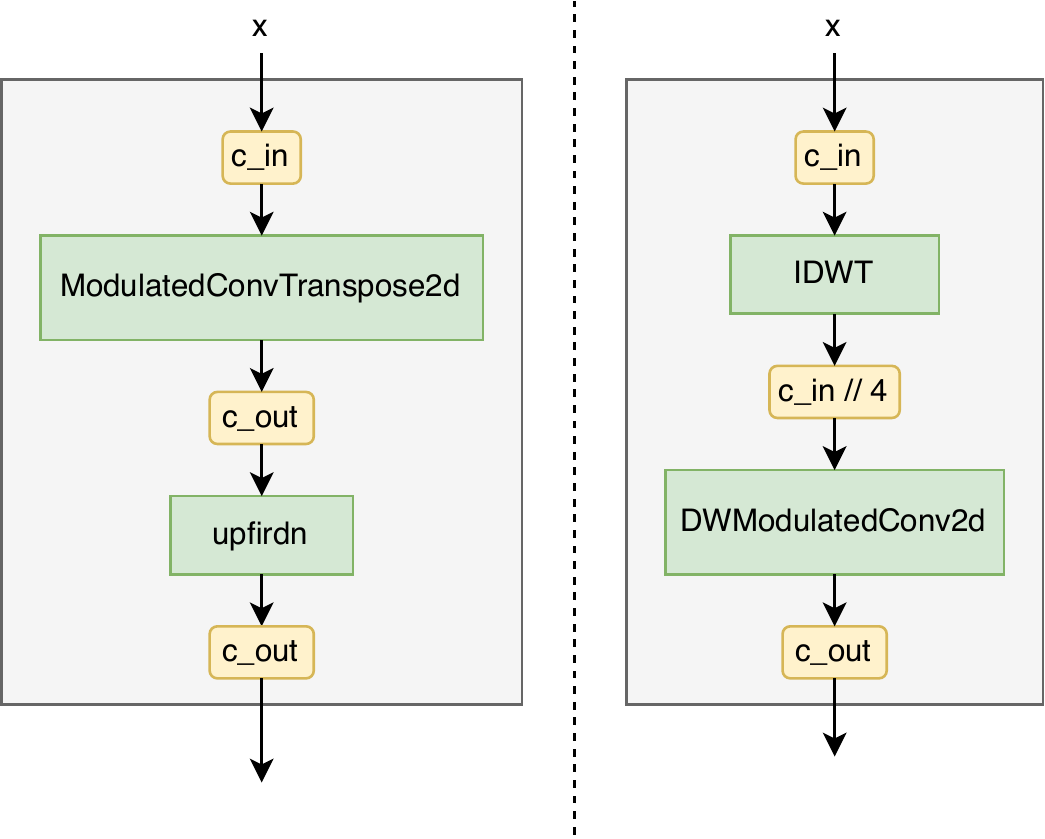}
    \end{center}
    \caption{(left) StyleGAN2 upscale block. (right) MobileStyleGAN upscale block. Yellow rectangles on arrows show the number of channels in related feature maps.}
    \label{fig:upscale_blosk}
\end{figure}

While the StyleGAN2 building block uses ConvTranspose (left panel of Figure \ref{fig:upscale_blosk}) to upscale input feature maps, following Section \ref{image_representation_revisited}, we use IDWT as an upscale function in the building block of MobileStyleGAN (right panel of Figure \ref{fig:upscale_blosk}). Since IDWT does not include trainable parameters, we add extra depthwise separable modulated convolution after the IDWT layer.

\begin{figure}[t]
    \begin{center}
    \includegraphics[width=1.0\linewidth]{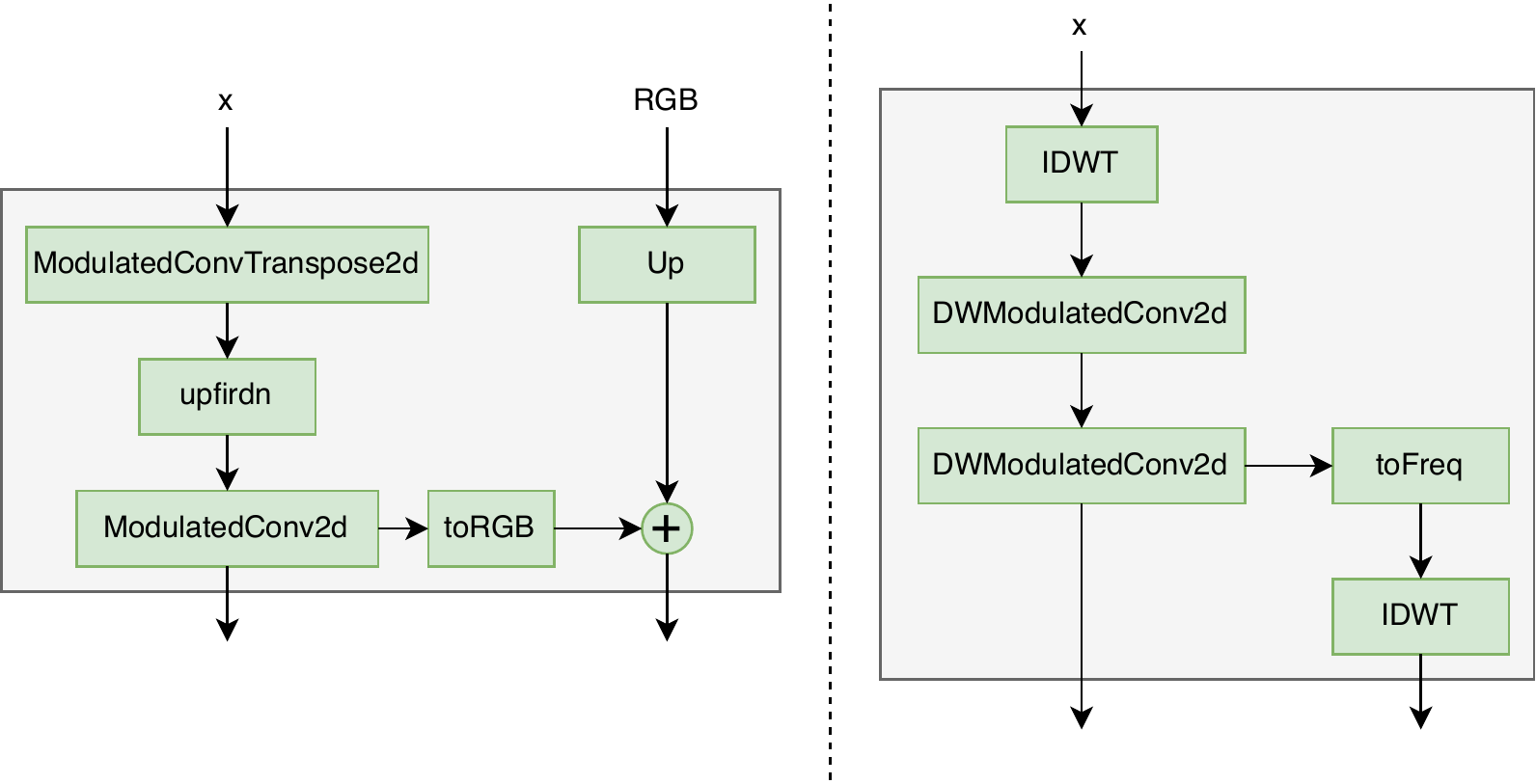}
    \end{center}
    \caption{(left) StyleGAN2 building block. (right) MobileStyleGAN building block.}
    \label{fig:building_block}
\end{figure}

The complete building block structures for StyleGAN2 and MobileStyleGAN are shown in Figure \ref{fig:building_block}.

\section{Training framework}

\begin{figure*}
    \begin{center}
    \includegraphics[width=1.0\linewidth]{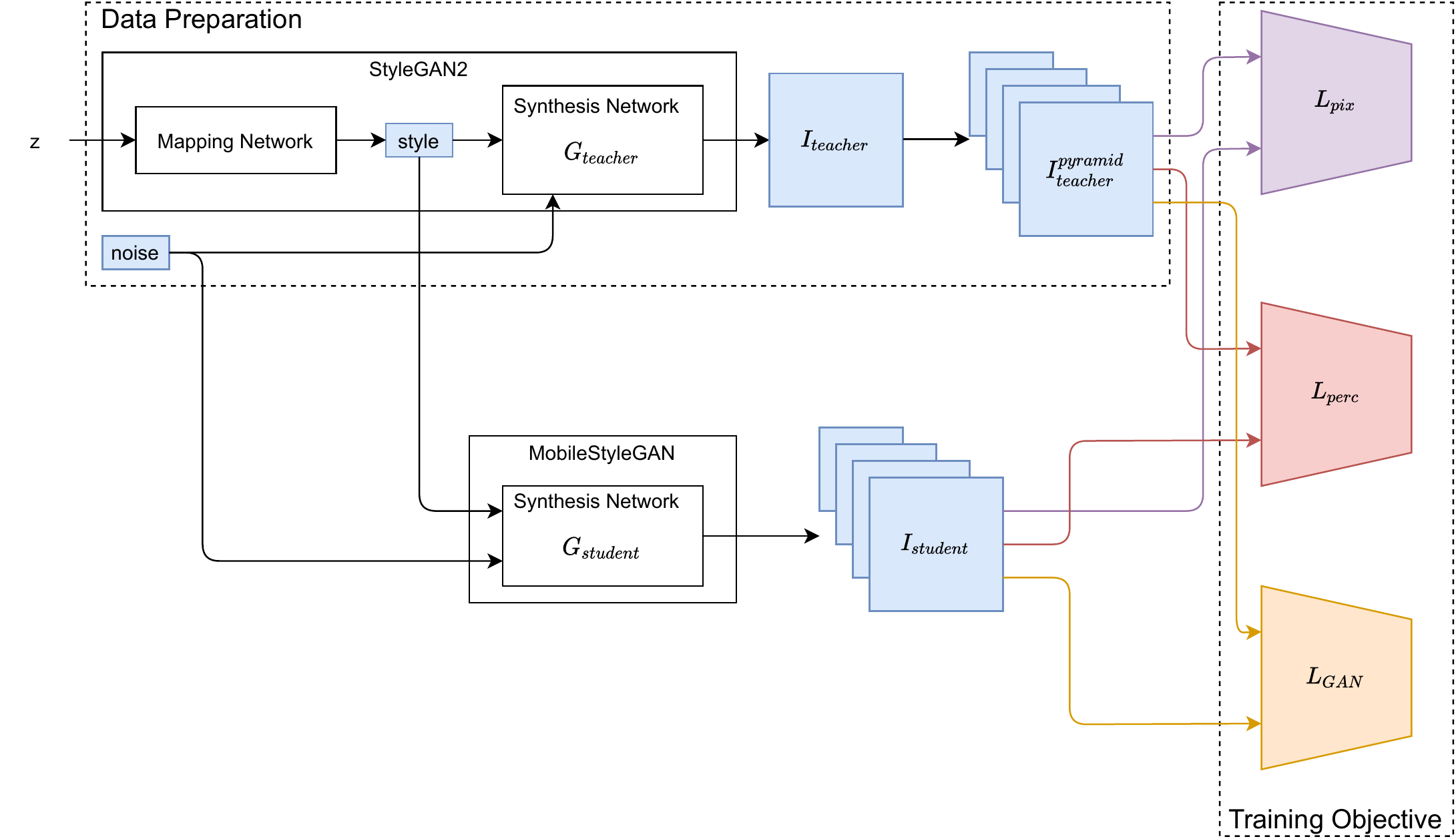}
    \end{center}
    \caption{Training framework.}
    \label{fig:distillation_pipeline}
\end{figure*}

As in previous works \cite{chang2020tinygan, li2020gan}, our training framework is based on the knowledge distillation technique \cite{hinton2015distilling}. Given StyleGAN2 \cite{Karras2019stylegan2} as the teacher network, we train MobileStyleGAN to mimic its functionality. The overall framework is illustrated in Figure \ref{fig:distillation_pipeline}. In this section, we discuss the main parts of our training framework.

\subsection{Data preparation}
Given the original StyleGAN2 generator, we can transform unpaired learning into the paired setting. To do so, we prepare triplet data $\{style,\ noise,\ I_{teacher}\}$, where $style$ is the output of the mapping network for given noise vector z, $noise$ is the noise shared between the teacher and student networks, and $I_{teacher}$ is the output of the teacher network for the given style.

As described in Section  \ref{progressive_growing_revisited}, each block of MobileStyleGAN predicts the output image for its spatial size. Therefore, instead of $I_{teacher}$, we use $I^{pyramid}_{teacher}$ as a ground-truth. $I^{pyramid}_{teacher}$ is the image pyramid built from $I_{teacher}$. Accordingly, our trained data are represented as the triplet data $\{style,\ noise,\ I^{pyramid}_{teacher}\}$.

To prevent overfitting, we do not use a preprocessed dataset. Instead, we generate data on the fly during a learning procedure.

To decrease memory consumption during the learning procedure, we use only artificial samples generated by StyleGAN2 and no real data.

\subsection{Training objective}
We now elaborate on the proposed objectives of the knowledge distillation.

\begin{figure}[t]
\begin{center}
   \includegraphics[width=1.0\linewidth]{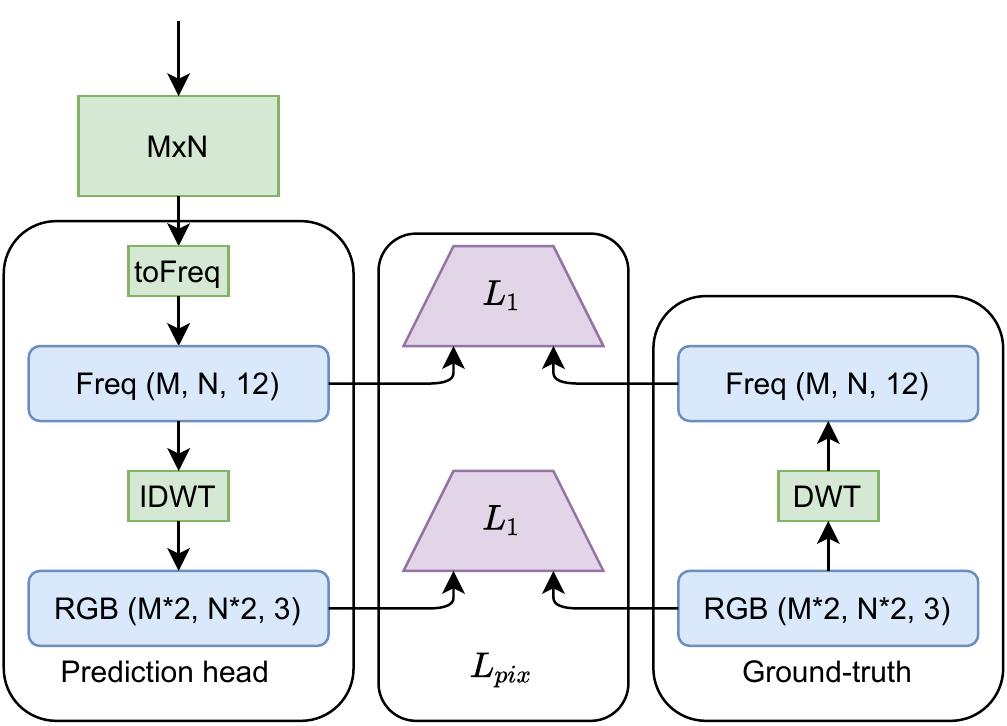}
\end{center}
   \caption{Pixel-Level Distillation Loss.}
\label{fig:pixel_level_loss}
\end{figure}

\textbf{Pixel-Level Distillation Loss (Figure \ref{fig:pixel_level_loss}).} Since MobileStyleGAN aims to predict the target image in the wavelet domain, the naive method to mimic the functionality of StyleGAN2 minimizes the pixel-level distance between the wavelet transform of the image generated by StyleGAN2 and the output of MobileStyleGAN. Also, we added a regularization term that minimizes the pixel-level distance between our ground truth and the predicted image in the pixel-based domain. We found that this term allows us to train different frequencies in concert with each other. As described in Section \ref{progressive_growing_revisited}, our network predicts output images at each spatial size. Accordingly, pixel-based distillation loss was applied at each scale.

Formally, let
\begin{equation}
    \begin{array}{l}
    L_{pix}(F_{s}, I_{t}) = \sum_{i}{(|| F^{i}_{s} - DWT(I^{i}_{t}) ||_1 + || IDWT(F^{i}_{s}) - I^{i}_{t} ||_1)}
    \end{array}
\end{equation}
where $F^{i}_{s}$ is the image in the wavelet domain predicted by the i-th block of the student network, $I^{i}_{t}$ is the ground-truth image in the pixel-based domain in the corresponding spatial size.

\textbf{Perceptual Loss.} The pixel-level loss described previously does not capture perceptual differences between output and ground-truth images. To solve this problem, we use perceptual loss as an objective. Our perceptual loss is based on VGG16 features and implemented as described in \cite{johnson2016perceptual}. We applied perceptual loss only to the output image generated by MobileStyleGAN.

Formally, let
\begin{equation}
    \begin{array}{l}
    L_{perc}(I_{s}, I_{t}) = \sum_{l}{|| VGG16(I^{256\mathbf{x}256}_{s})_{l} - VGG16(I^{256\mathbf{x}256}_{t})_{l} ||_2}
    \end{array}
\end{equation}
where $VGG16(...)_{l}$ are intermediate features from the corresponding layer $ l \in [relu1\_2,\ relu2\_2,\ relu3\_3,\ relu4\_3]$ of the VGG16, $I^{256\mathbf{x}256}_{s}$ is the output image predicted by the student network (resized to 256x256), where $I^{256\mathbf{x}256}_{t}$ is the output image predicted by the teacher network (resized to 256x256).

\textbf{GAN Loss.} Using only pixel-level and perceptual losses leads to generation of blurred images. To sharpen the generated images, we incorporate a discriminator network in our pipeline. We adopt the GAN loss for the generator:
\begin{equation}
    \begin{array}{l}
    L_{g}(style, noise) = f(-D_{T}(G_{student}(style, noise)))
    \end{array}
\end{equation}
and for the discriminator network:
\begin{equation}
    \begin{array}{l}
    \begin{aligned}
    L_{d}(style, noise) = {} & f(- D_{T}(G_{teacher}(style, noise))) \\
                           & + f(D_{T}(G_{student}(style, noise))) \\
                           & + \frac{\gamma}{2} \mu(||\nabla D_{T}(G_{teacher}(...)|| ^ 2)
    \end{aligned}
    \end{array}
\end{equation}
where $f(t) = - \log(1 + \exp(-t))$ is the softplus function, $D_{T}(...)$ is the discriminator network with differentiable augmentations, $G_{student}(...)$/$G_{teacher}(...)$ is the student/teacher network, and (style, noise) is the input paired data. The discriminator has the same topology as in \cite{karras2019stylebased}.

We found that it is hard to balance capacity between generator and discriminator networks when we compress the generator network. So, R1 regularization is an important part of GAN Loss for distillation pipelines, that allow for recouping capacity disbalance between generator and discriminator.

\textbf{Full objective.} Finally, we define the full objective as:
\begin{equation}
    \begin{array}{l}
    L_{student} = \lambda_1 * L_{pix} +\lambda_2 * L_{perc} + \lambda_3 * L_{g} \\
    L_{discriminator} = L_{d}
    \end{array}
\end{equation}
where hyperparameters $\lambda_1,\ \lambda_2,\ \lambda_3$ control the importance of each term.

\section{Experiments}

To train our MobileStyleGAN network we use the StyleGAN2 teacher network trained on the FFHQ \cite{karras2019stylebased} dataset.

\subsection{Training}
MobileStyleGAN is trained using the Adam algorithm \cite{kingma2017adam} with $\beta1=0.9,\ \beta2=0.999$. The generator and discriminator learning rates are both set to constant 5e-4. We use affine transforms and cutout as differentiable augmentations at the input of the discriminator. Hyperparameters of the objective function set are fixed at $\lambda_1=1.0,\ \lambda_2=1.0,\ \lambda_3=0.1$. We update the generator and discriminator at each optimization step. Training takes about three days on 4 x NVIDIA 2080Ti GPU with batch\_size=8.

\subsection{Results}
Table \ref{table:results} shows the result of our evaluation of MobileStyleGAN on the FFHQ datase. We compare the number of parameters, the computational cost and the Frechet inception distance (FID) \cite{heusel2018gans} of MobileStyleGAN and the teacher network (StyleGAN2).

\begin{table}
\begin{center}
\begin{tabular}{|c|c|c|c|}
\hline
Network & MParams & GMACs & FID \\
\hline\hline
StyleGAN2 & 28.27 & 143.15 & 2.84 \\
MobileStyleGAN & 8.01 & 15.09 & 7.75 \\
\hline
\end{tabular}
\end{center}
\caption{Main results. Comparison between StyleGAN and MobileStyleGAN for the FFHQ dataset, 1024x1024.}
\label{table:results}
\end{table}

Also, we evaluate inference time for StyleGAN2 and MobileStyleGAN on CPU. We use a laptop with Intel(R) Core(TM) i5-8279U CPU for our experiments. Table \ref{table:inference} shows the result of inference time evaluation. As we noticed previously, some architectural tricks that we used in MobileStyleGAN make available graph optimizations such as constant folding, etc. We get better performance when we use special inference engines such as OpenVINO \cite{openvino} that automatically apply graph optimization.

\begin{table}
\begin{center}
\begin{tabular}{|c|c|c|}
\hline
Network & Engine & Time (sec.) \\
\hline\hline
StyleGAN2 & PyTorch & 4.3 \\
MobileStyleGAN & PyTorch & 1.2 \\
MobileStyleGAN & OpenVINO & 0.16 \\
\hline
\end{tabular}
\end{center}
\caption{Inference time on CPU (Intel(R) Core(TM) i5-8279U).}
\label{table:inference}
\end{table}

\section{Conclusion}

In this work, we approached the problem of high-fidelity image synthesis, suitable for deployment on edge devices. We proposed the new style-based lightweight generative network and training pipeline based on knowledge distillation. We have made our training code publicly available (\url{https://github.com/bes-dev/MobileStyleGAN.pytorch}), along with a simple python library for fast random face synthesis on the CPU (\url{https://github.com/bes-dev/random_face}). The accompanying videos can be found on YouTube (\url{https://www.youtube.com/playlist?list=PLstKhmdpWBtwsvq_27ALmPbf_mBLmk0uI}).

Some techniques may further improve performance and accuracy, such as quantization and pruning. We left them for future research.

{\small
\bibliographystyle{ieee_fullname}
\bibliography{egbib}
}

\end{document}